\pdfoutput=1

\documentclass[11pt]{article}

\usepackage[final]{emnlp2021}

\usepackage{times}
\usepackage{latexsym}

\usepackage[T1]{fontenc}

\usepackage[utf8]{inputenc}

\usepackage{microtype}

\usepackage{multirow}
\usepackage{CJKutf8}
\usepackage{amsmath}
\usepackage{xspace}
\usepackage[switch]{lineno}
\usepackage{enumitem}
\usepackage{amsfonts}
\usepackage{booktabs}
\usepackage{tabulary}
\usepackage{graphicx}
\usepackage{ulem}

\title{Secoco: Self-Correcting Encoding for Neural Machine Translation}

\author{
    Tao Wang\textsuperscript{1}, Chengqi Zhao\textsuperscript{1}, Mingxuan Wang\textsuperscript{1}, Lei Li\textsuperscript{1}, Hang Li\textsuperscript{1}, Deyi Xiong\textsuperscript{2} \\
    \textsuperscript{1}ByteDance AI Lab \\
    \textsuperscript{2}College of Intelligence and Computing, Tianjin University, Tianjin, China \\
    \texttt{\{wangtao.960826, zhaochengqi.d, wangmingxuan.89\}@bytedance.com} \\
    \texttt{\{lilei.02, lihang.lh\}@bytedance.com} \\
    \texttt{dyxiong@tju.edu.cn}
}

\newcommand{\method}{{Secoco}\xspace}
\newcommand{\mbase}{\textsc{Base}\xspace}
\newcommand{\mrepair}{\textsc{Repair}\xspace}
\newcommand{\mrec}{\textsc{Reconstruction}\xspace}

\begin{document}
\begin{CJK}{UTF8}{gbsn}
\maketitle

\begin{abstract}
%

This paper presents \textbf{Se}lf-\textbf{co}rrecting En\textbf{co}ding (\method), a framework that effectively deals with input noise for robust neural machine translation by introducing self-correcting predictors.  
Different from previous robust approaches, \method 
enables NMT to explicitly correct noisy inputs and delete specific errors simultaneously with the translation decoding process. \method is able to achieve significant improvements over strong baselines on two real-world test sets and a benchmark WMT dataset  with good interpretability.
We will make our code and dataset publicly available soon.

\end{abstract}

\section{Introduction}
\label{sec:intro}


Neural machine translation (NMT) has witnessed remarkable progress in recent years~\cite{bahdanau2014neural,vaswani2017attention}.
Most previous works show promising results on clean datasets, such as WMT News Translation Shared Tasks \cite{barrault2020proceedings}.  However, inputs in real-world scenarios are usually with a wide variety of noises, which poses a significant challenge to NMT.

\begin{figure}[t]
\centering
\includegraphics[scale=1.0]{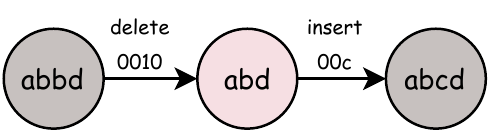}
\caption{An example of the correcting and operation generation process. 
Assume we want to correct synthesized noisy sequence ``abbd" to its correct sequence ``abcd".
We can apply a deletion ``b" to the third position (0010) and an insertion ``c" to the third position (00c).
(``abbd",``0010") and (``abd",``00c") can be regarded as training examples.}
\label{fig:approach:dataflow}
\end{figure}

In order to mitigate this issue, we propose to build a noise-tolerant NMT model with a \textbf{Se}lf-\textbf{co}rrecting En\textbf{co}ding (\method) framework that explicitly models the error-correcting process as a sequence of operations: deletion and insertion.
Figure \ref{fig:approach:dataflow} demonstrates a simple correcting process that transforms a noisy sequence "abbd" into its correct sequence "abcd" via a deletion and inserting operation. In order to learn desirable operations for noise correction given noisy inputs, we propose a insertion predictor and deletion predictor that predict appropriate deletion and insertion operations respectively. The two predictors work alternatively step by step to collectively transform a noisy input sequence into a clean sequence. 

For training the two predictors, we collect  a list of pairs (source sequence, operation sequence) (e.g.,  (“abbd",“0010") shown in Figure 1) from original training data by randomly deleting or inserting tokens to original clean sequences. With these collected training instances, we optimize the insertion and deletion predictors as well as NMT
simultaneously in a multi-task learning way. 

For inference, we propose two different variants for \method depending on the decoding modes.
The first variant is an end-to-end approach like normal NMT decoding where the encoder is implicitly trained with self-correcting information. In this setting, we only predict operations during training and the encoder can have this kind of knowledge.
The other variant is iterative editing, which corrects the input gradually and performs translation after the input is unchanged.

\begin{figure*}[t]
\centering
\includegraphics[scale=1.0]{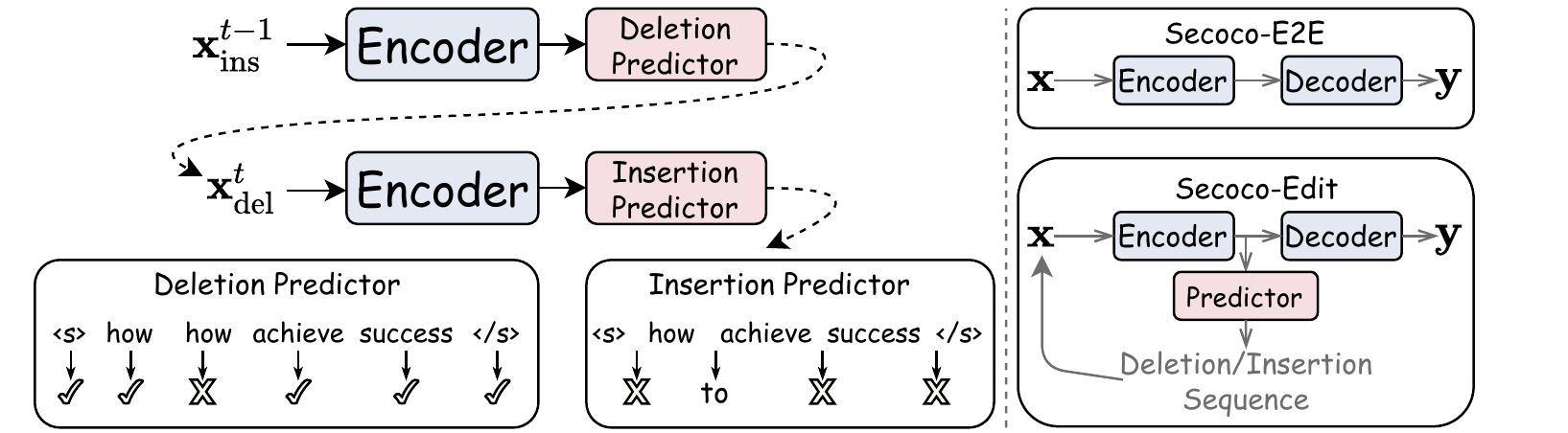}
\caption{Diagram of the proposed \method. The left part is the illustration of self-correcting encoding. It contains a deletion predictor (Eq. \ref{eq:approach:delete}) and an insertion predictor (Eq. \ref{eq:approach:insert}).
We omit the translation part here due to the space limit.
The right part shows the decoding modes.}
\label{fig:approach:illustration}
\end{figure*}

Compared with previous approaches, \method has two advantages. 
First, \method introduces a more explicit and direct
 way to model the noise correcting process.
Second, \method enables an interpretable translation process. With the predicted operation sequence, it is easy to understand how the noisy input is corrected.
We conduct experiments on three test sets, including Dialogue, Speech, and WMT14 En-De tasks. The results show that \method outperforms the baseline by $+1.6$ BLEU.


\section{Approach}
\label{sec:approach}

Our approach is illustrated in Figure 2. 
The left part of Figure \ref{fig:approach:illustration} demonstrates the encoding module of \method.
The only difference of \method from standard translation models is the two correcting operation predictors, which generate the operation sequence based on the encoder representation of an input text. 
The deletion predictor decides which word to be deleted while the insertion predictor decides which word to be inserted into which position. The combination of these two operations is able to simulate arbitrary complex correcting operations \cite{DBLP:conf/nips/GuWZ19}.

We illustrate the training data synthesizing process for the two predictors in Figure \ref{fig:approach:dataflow}.
It is worth noting that for correcting that contains several iterations of editing (i.e., deletion or insertion), we sample only one iteration from it.

\subsection{Self-Correcting Encoding}
\method iteratively applies deletion and insertion operations to obtain a clean source sentence from a noisy input source sentence. Formally, given a source sentence $\mathbf{x}$, we introduce $\mathbf{x}_{\rm{del}}^t$ and $\mathbf{x}_{\rm{ins}}^t$ as the edited sentences at the $t$-th iteration after the deletion and insertion operation is respectively performed. 
As illustrated in the left part in Figure~\ref{fig:approach:illustration}, the deletion predictor decides whether to delete (1) or keep unchanged (0) at position $i$:
\begin{equation}
    p(c^t_i|\mathbf{x}_{\rm{ins}}^{t-1}) = {\rm sigmoid}(h_{{\rm{ins}},i}^{t-1}\, W)
\label{eq:approach:delete}
\end{equation}
where $c_i^t \in \{0, 1\}$, $W\in \mathbb{R}^{d\times 2}$ and $h_{{\rm{ins}}, *}^{t-1} \in \mathbb{R}^{1\times d}$ is the encoded source representation after $(t-1)$ iterations.

Similarly, the insertion predictor considers the positions between each pair of neighboring words, and predicts a word to be inserted at position $j$:
\begin{equation}
    p(w^t_j| \mathbf{x}_{\rm{del}}^t) = {\rm softmax}([h_{{\rm{del}},j}^t;h_{{\rm{del}},j+1}^t] \, Z)
\label{eq:approach:insert}
\end{equation}
where $Z \in \mathbb{R}^{2d\times (|V|+1)}$ and $h_{{\rm{del}},*}^t$ is the encoded representation after deletion at the $t$-th iteration. Here, 
$|V|$ is the source vocabulary size and we append an empty token into the vocabulary, denoting no insertion operation at that position.

Although the iterative editing process relies heavily on previous operations for both the prediction of deletion and insertion, the two predictors are independently trained for simplicity. The training data generated in advance is used to train both the deletion and insertion predictors simultaneously.

\subsection{Training Objectives}
We build the \method based on the encoder-decoder framework. Given a source sentence $\mathbf{x}$ and its target translation $\mathbf{y}=\{y_1, ..., y_m\}$, NMT directly models the conditional probability of the target sentence over the source sentence:
\begin{equation}
    p\left(\mathbf{y}|\mathbf{x} \right) = \prod_{i=1}^m p\left(y_i|\mathbf{x},y_{<i} \right)
\end{equation}
As for deletion and insertion predictors, assume we have the supervision $\{\mathbf{c}^t, \mathbf{w}^t\}$ for each iteration $t \in {1,...,T}$.
We can jointly train the above three tasks, and the training objective is to maximize the overall log-likelihood:
\begin{equation}
\begin{aligned}
    \log p(\mathbf{y}|\mathbf{x}) + \sum_{t=1}^T \left(
    \log p(\mathbf{c}^t|\mathbf{x}_{\rm{ins}}^{t-1}) 
    + \log p(\mathbf{w}^t|\mathbf{x}_{\rm{del}}^t)\right)
\end{aligned}
\label{eq:approach:training}
\end{equation}
where $T$ is set to 1 when we only sample one iteration of editing during training.
\begin{table}[]
\centering
\small
\begin{tabulary}{1.\linewidth}{p{28pt}p{12pt}p{77pt}l}
\toprule
Test set & Size & Noise Types & Edits \\
\midrule
\multirow{3}{*}{Dialogue} & \multirow{3}{*}{1,931} & dropped pronoun & delete \\
& & dropped punctuation & delete \\
& & typos & delete+insert\\
\midrule
\multirow{2}{*}{Speech} & \multirow{2}{*}{1,389} & spoken words & insert \\
& & wrong punctuation & delete+insert \\
\midrule
\multirow{3}{*}{WMT} & \multirow{3}{*}{3,000} & random insertion & insert\\
& & random deletion & delete\\
& & repeated words & insert \\
\bottomrule
\end{tabulary}
\caption{Details of the three test sets.}
\label{table:exp:perturb}
\end{table}

\subsection{Decoding Modes}
\label{sec:approach:modes}
During inference, we can either use the encoder-decoder model only (\method-E2E) that is trained with the two predictors simultaneously or translate the edited sentence after iteratively applying deletion and insertion operations (\method-Edit), as illustrated in the right part of Figure \ref{fig:approach:illustration}.


In general, \method-E2E provides better robustness without sacrificing decoding speed. For \method-Edit, iterative editing enables better interpretability.
Detailed editing operations provide a different perspective on how the model resists noise.

\section{Experiments}
\label{sec:exp}
\begin{table*}[]
\centering
\small
\begin{tabulary}{1.\linewidth}{lccccccccc}
\toprule
\multirow{2}{*}{Methods} & \multicolumn{2}{c}{Dialogue} & \multicolumn{2}{c}{Speech} & \multicolumn{2}{c}{WMT En-De} & \multicolumn{2}{c}{AVG} & Latency \\
& BLEU & $\triangle$ & BLEU & $\triangle$ & BLEU & $\triangle$ & BLEU & $\triangle$ & (ms/sent)\\
\midrule
\mbase & 31.8 & N/A & 11.1 & N/A & 24.5 & N/A & 22.5 & N/A & 22\\
\mbase+synthetic & 32.6 & +0.8 & 11.7 & +0.6 & 24.8 & +0.3 & 23.0 & +0.5 & 21\\
\mrepair & 33.2 & +1.4 & 11.4 & +0.3 & 25.0 & +0.5 & 23.2 & +0.7 & 36\\
\mrec & 33.7 & +1.9 & 11.8 & +0.7 & 24.6 & +0.1 & 23.4 & +0.9 & 21\\
\midrule
\method-Edit & 34.1 & +2.3 & 12.3 & +1.2 & \textbf{25.2} & \textbf{+0.7} & 23.9 & +1.4 & 24\\
\method-E2E & \textbf{34.8} & \textbf{+3.0} & \textbf{12.4} & \textbf{+1.3} & 25.1 & +0.6 & \textbf{24.1} & \textbf{+1.6} & 22\\
\bottomrule
\end{tabulary}
\caption{Experiment results on the Dialogue, Speech and WMT En-De translation test set. We evaluate the average latency over the three test sets.}
\label{table:exp:main_results}
\end{table*}

\subsection{Data}
We conducted our experiments on three test sets, including Dialogue, Speech, and WMT14 En-De, to examine the effectiveness of \method.

Dialogue is a real-world Chinese-English dialogue test set constructed based on TV drama subtitles, which contains three types of natural noises ~\cite{wang2021autocorrect}.
Speech is an in-house Chinese-English speech translation test set which contains various noise from ASR.
To evaluate \method on different language pairs, we also used WMT14 En-De test sets to build a  noisy test set with random deletion and insertion operations.
Table \ref{table:exp:perturb} shows the details of the three test sets.

For Chinese-English translation, we used  WMT2020 Chinese-English data\footnote{http://www.statmt.org/wmt20/translation-task.html} (48M) for Dialogue,
and CCMT\footnote{This corpus is a part of WMT2020.} (9M) for Speech.
For WMT En-De, we adopted the widely-used WMT14 training data\footnote{http://www.statmt.org/wmt14/translation-task.html} (4.5M).
We synthesized corresponding noisy data according to the noise types of the corresponding test set.
The test sets and codes for synthesizing noisy data used in our experiments will be released.

\subsection{Baselines}
We compared our method against the following three baseline systems.

\noindent\textbf{\mbase}
One widely-used way to achieve NMT robustness is to mix raw clean data with noisy data to train NMT models. We refer to models trained with/without synthetic data as \mbase/\mbase {+synthetic}.

\noindent\textbf{\mrepair}
To deal with noisy inputs, one might train a repair model to transform noisy inputs into clean inputs that a normally trained translation model can deal with. 
Both the repair and translation model are transformer-based models.
As a pipeline model (repairing before translating), \mrepair may suffer from error propagation.

\noindent\textbf{\mrec}
We follow \citet{zhou2019improving} to develop a multi-task based method to solve the robustness problem. We construct triples (clean input, noisy input, target translation), and introduce an additional decoder to obtain clean inputs from noisy inputs. This method enables NMT to transform a noisy input into a clean input and  pass this knowledge into the translation decoder.

\subsection{Settings}
In our studies, all translation models were Transformer-base.
They were trained with a batch size of 32,000 tokens.
The beam size was set to 5 during decoding.
We used byte pair encoding compression algorithm (BPE) \cite{sennrich2016neural} to process all these data and restricted merge operations to a maximum of 30k separately.
For evaluation, we used the standard Sacrebleu ~\cite{post2018call} to calculate BLEU-4.
All models were implemented based on Fairseq \cite{ott2019fairseq}.

\subsection{Results}
Table \ref{table:exp:main_results} shows the translation results on Dialogue, Speech and WMT En-De.
Clearly, all competitors substantially improve the baseline model in terms of BLEU.
\method achieves the best performance on all three test sets, gaining improvements of 2.2, 0.7, and 0.4 BLEU-4 points over \mbase{+synthetic} respectively. 
The improvements suggest the effectiveness of self-correcting encoding.

It is worth noticing that the BLEU scores here are results of noisy test sets, so they are certainly lower than the results without noise.

Among these test sets, Dialogue is much more noisy and informal than the other two test sets. 
\method-E2E achieves a BLEU score of 34.8, which is even 3 BLEU points higher than the baseline. 
Speech is very challenging  and contains many errors introduced by ASR. 
The best BLEU score of Speech is only 12.4, achieved by \method-E2E. 
We have additional two interesting findings.
First, the performance of \method-E2E and \method-Edit is very close. Therefore, it is better to use \method-E2E for its simplification and efficiency. 
Second, \method is more effective on the real-world test sets, showing its potential in real-world application.    



\begin{table}[t]
\centering
\small
\begin{tabulary}{1.\linewidth}{ccl}
\toprule
Iteration & Edition & Sentence\\
\midrule
0 & & We has things to to do today \\
\midrule
\multirow{2}{*}{1} & delete & We \sout{has} things to \sout{to} do today\\  
& insert & We \uline{have} things to do today\\
\midrule
\multirow{2}{*}{2} & no delete & \\
& insert & We have things to do today \uline{.}\\
\bottomrule
\end{tabulary}
\caption{An example of the editing process using \method-Edit. The raw sentence is ``We have things to do today.". \sout{word} is to be deleted while \uline{word} is to be inserted.}
\label{table:analysis:edit}
\end{table}
\normalem


\subsection{Iterative Editing}
As described in Section \ref{sec:approach:modes}, we iteratively edit the input until the input is unchanged and then translate it. We present an examples in Table \ref{table:analysis:edit}. We can see that multiple deletions can be parallel, and the same is true for insertions. Because we try to make editing sequences as short as possible during the training process, we usually need only 1 to 3 iterations during inference. We get an average iteration number of 2.3 on our three test sets.


\section{Related Work}
\label{sec:relatedwork}
Approaches to the robustness of NMT can be roughly divided into three categories.
In the first research line, adversarial examples are generated with back-or white-box methods. The generated adversarial examples are then used to combine with original training data for adversarial training~\cite{ebrahimi2018adversarial, chaturvedi2019exploring, cheng2019robust, michel2019evaluation, zhao2018generating, cheng2020advaug}. 

In the second strand, a wide variety of methods have been proposed to deal with noise in training data~\cite{schwenk2018filtering, guo2018effective, xu2017zipporah, koehn2018findings, van2017dynamic, wang2019target, wang2018dynamic, wang2018denoising, wang2019dynamically}.

Finally, efforts have been also explored to directly cope with naturally occurring noise in texts, which are closely related to our work. 
\citet{heigold2018robust, belinkov2018synthetic, levy2019training} focus on word spelling errors. 
\citet{sperber2017toward, liu2019robust} study translation problems caused by speech recognition.
\citet{vaibhav2019improving} introduce back-translation to generate more natural synthetic data, and employ extra tags to distinguish synthetic data from raw data.
\citet{zhou2019improving} propose a reconstruction method based on one encoder and two decoders architecture to deal with natural noise for NMT.
Different from ours, most of these works use the synthetic data in a coarse-grained and implicit way (i.e. simply combining the synthetic and raw data).

\section{Conclusions}
\label{sec:conclusions}
In this paper, we have presented a framework \method to build a noise-tolerant NMT model with self-correcting capability.
With the proposed \method-E2E and \method-Edit methods, \method exhibits both efficiency and interpretability.
Experiments and analysis on the three test sets demonstrate that the proposed \method is able to improve the quality of NMT in translating noisy inputs, and make better use of synthetic data.

\bibliographystyle{acl_natbib}
\bibliography{anthology,custom}


\end{CJK}
\end{document}